\documentclass[letterpaper, 10 pt, conference]{ieeeconf}  

\IEEEoverridecommandlockouts                              
\usepackage[utf8]{inputenc}
\usepackage[T1]{fontenc}

\usepackage{amsmath}
\usepackage{amssymb}
\usepackage{graphicx}
\usepackage{booktabs}
\usepackage{multirow}
\usepackage{makecell}
\usepackage{array}
\usepackage{pgfplots}
\pgfplotsset{compat=1.18}
\usetikzlibrary{patterns}
\usepackage{url}
\usepackage{capt-of}

\title{\LARGE \bf
Decoupled Asynchronous System for Humanoids (DASH): Latency-Agnostic Scaling of Vision-Language-Action Models for Humanoid Manipulation 
}

\author{
Bingzhang Wang$^{1,2,*}$,
Nathan Zhao$^{1}$,
Jialeng Ni$^{1}$,
Jiayin Zheng$^{1}$,
Yin Zhang$^{1}$,
Kehua Chen$^{2}$,
and Jian Yao$^{1}$%
\thanks{$^{1}$XPENG Robotics, Santa Clara, CA, USA.}%
\thanks{$^{2}$University of Washington, Seattle, WA, USA.}%
\thanks{$^{*}$Corresponding author: Bingzhang Wang.}%
}

\IEEEaftertitletext{%
  \vspace{0.4\baselineskip}%
  \noindent\begin{minipage}{\textwidth}\centering
    \includegraphics[width=\textwidth]{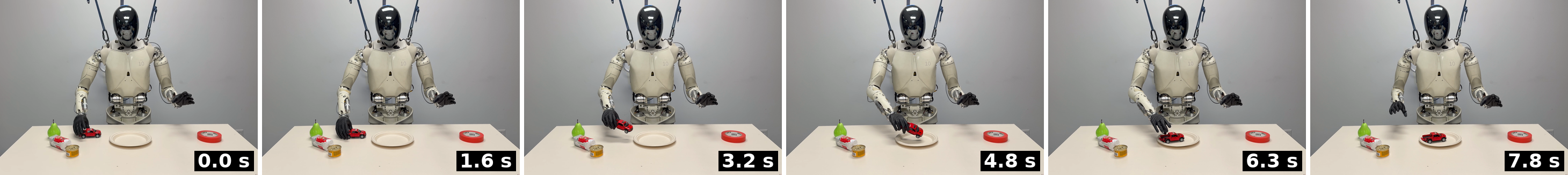}
    \captionof{figure}{DASH on a real humanoid with dexterous hands executing ``pick up the toy car with the right hand and put it to the plate'': six frames at equal intervals of wall-clock time (time stamps in the corners).}\label{fig:teaser}
  \end{minipage}\vspace{0.8\baselineskip}}

\begin{document}

\maketitle
\thispagestyle{empty}
\pagestyle{empty}

\begin{abstract}
Vision-language-action (VLA) models pair a vision-language backbone with an action head, but in closed-loop control the backbone sits on the control-critical path: while it computes, the robot keeps moving, so each executed action is conditioned on an increasingly stale observation. A larger backbone therefore directly blocks reactivity, especially in humanoid dexterous manipulation. We present DASH, a Decoupled Asynchronous System for Humanoids that removes this coupling. DASH splits the policy into a slow vision-language model (System~2) and a fast flow-matching action head (System~1) that run as concurrent processes over the LCM communication protocol, so the control rate is set by the action head alone, independent of backbone size. The action head fuses stale semantics from a frozen backbone with fresh vision from its own ResNet encoder, desynchronized training exposes it to the staleness it meets at deployment, and an adaptive temporal aggregator aligns overlapping action chunks under variable inference latency. On an asynchronous RoboTwin benchmark that charges inference latency in simulated time, DASH holds $20$\,Hz control rate from 0.8B to 8B backbones while the synchronous rate falls from $9.7$ to $4.1$\,Hz, with higher success at every size. On a real humanoid with dexterous hands, DASH achieves clearly higher pick-and-place success than the synchronous pipeline.
\end{abstract}

\section{Introduction}


Vision-Language-Action (VLA) models have become a default recipe for generalist robot manipulation, pairing a pretrained vision-language backbone with an action head trained end-to-end on demonstrations~\cite{rt2,openvla,pi0,gr00tn1}.
Deploying them on a real humanoid (Fig.~\ref{fig:teaser}), however, is a fundamentally time-sensitive problem.
Unlike open-loop evaluation, where a policy is scored on a fixed dataset with no regard for inference cost, closed-loop control couples the model to a world that never stops moving: while the model computes the next actions, the robot is already acting and cannot pause to wait.
This mismatch is intrinsic to embodied systems, and it grows with model latency.

\begin{figure}[!t]
    \centering
    \includegraphics[width=\linewidth]{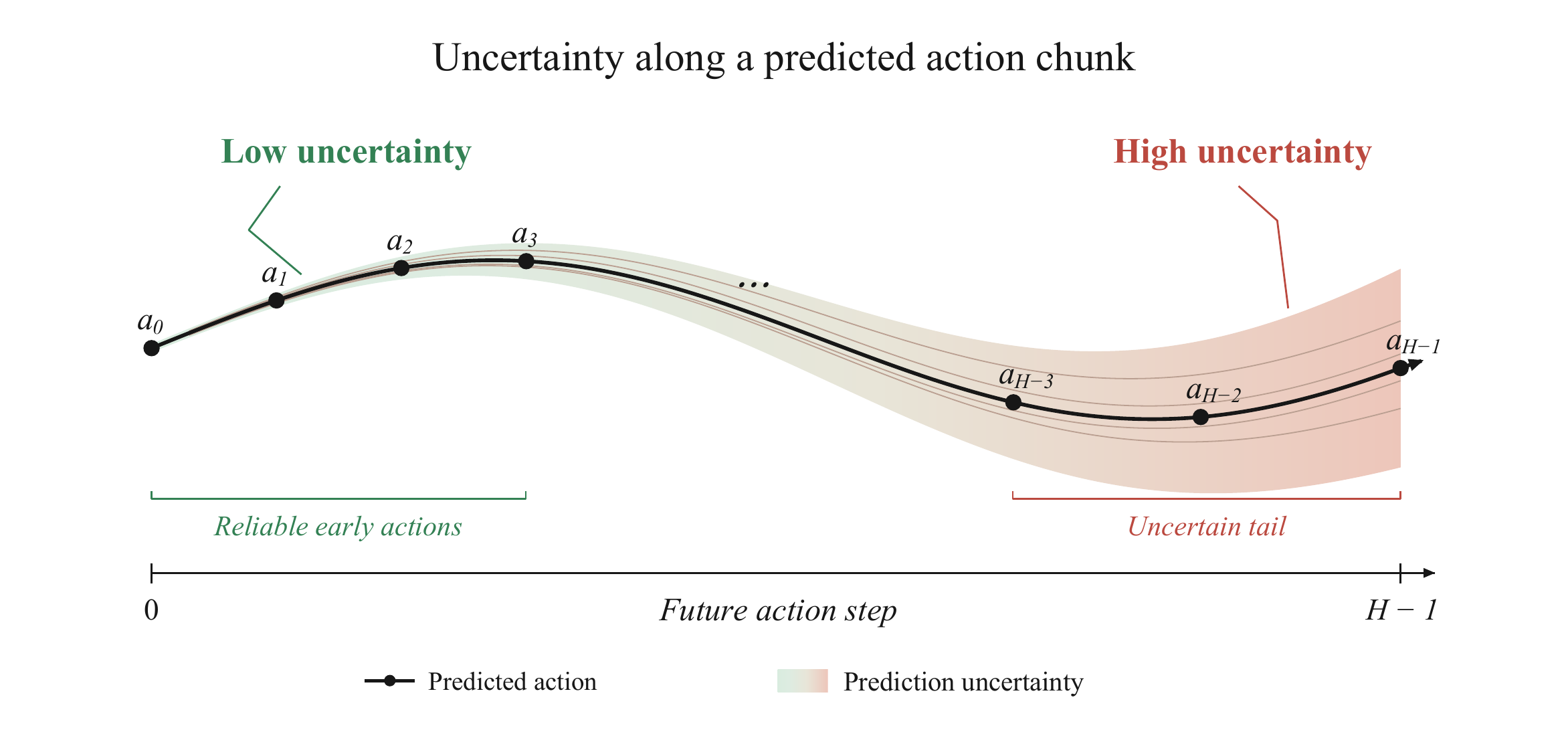}\\[1pt]
    {\small (a)}\\[5pt]
    \includegraphics[width=\linewidth]{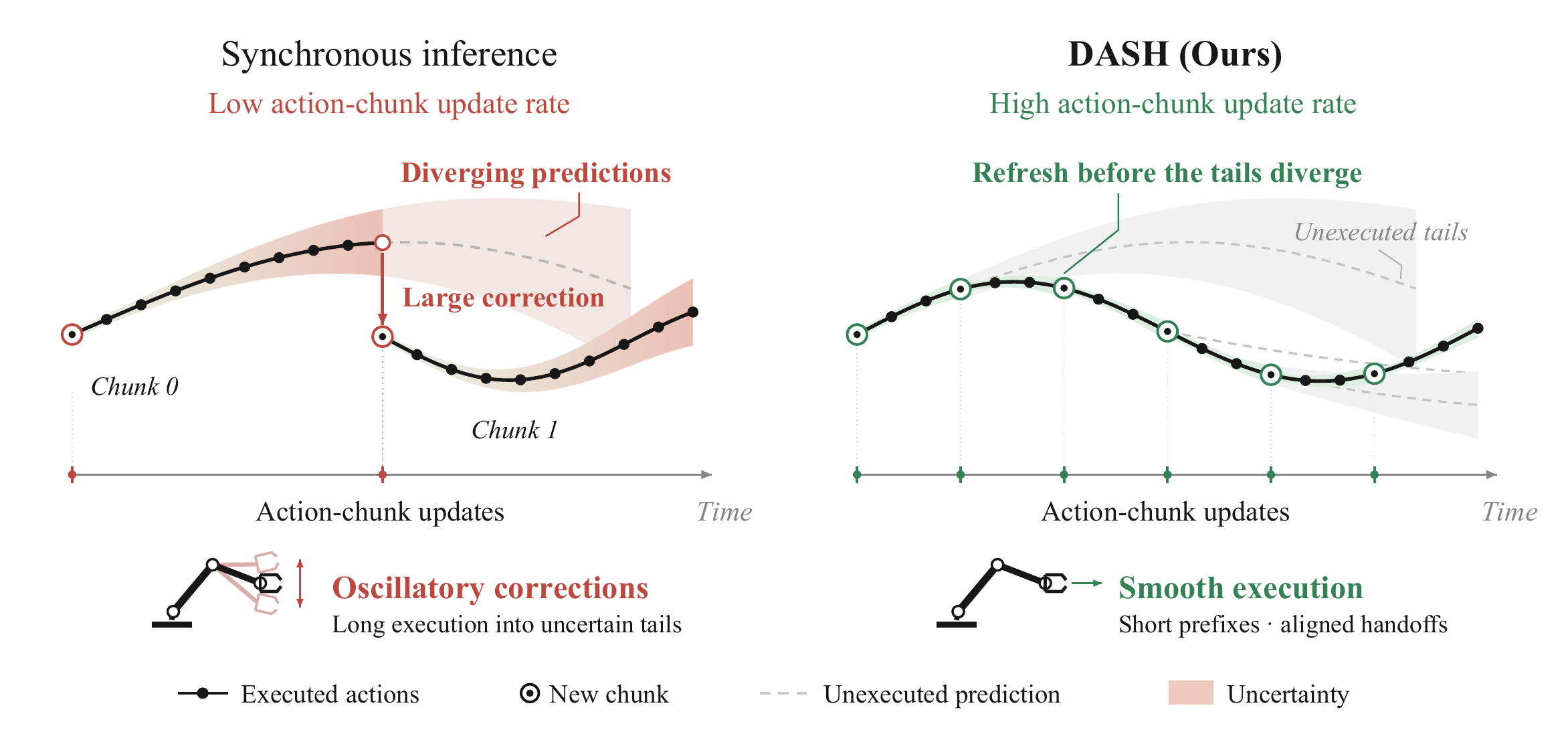}\\[1pt]
    {\small (b)}
    \caption{(a) Uncertainty grows along the prediction horizon: the early portion of an action chunk is reliable, while its distant tail admits increasingly different future actions. (b) Frequent chunk updates curb divergence: synchronous inference executes the long, uncertain tail of each chunk, so a new prediction can trigger a large correction; DASH refreshes chunks often and stays on the reliable early portion.}
    \label{fig:intro}
\end{figure}

Concretely, an action chunk computed from the state at time $0$ becomes available only after the inference time $L$, when the robot has already moved on, so executing it produces delayed, spatially misaligned actions.
This error is invisible to open-loop metrics such as validation loss, but in closed loop it surfaces as slow, hesitant, discontinuous, and even shaking manipulation, precisely the regime that matters for dexterous humanoid hands.

This temporal gap collides head-on with the scaling law that drives VLA progress.
Larger vision-language backbones generalize better and follow instructions more reliably~\cite{rt2}, but they are also slower, and higher latency degrades closed-loop control through exactly the staleness above.
Model size thus becomes a trade-off between semantic generalization and control reactivity, a trade-off that is especially punishing for humanoid manipulation, which demands both broad world understanding and high-frequency, contact-rich dexterity.


We argue that this trade-off is an artifact of synchronous execution rather than a law of nature.
A real robot is already asynchronous: sensing, high-level understanding, and low-level actuation run at different natural rates.
We therefore decouple the VLA into a slow semantic stream (the vision-language backbone, ``System~2'') and a fast reactive stream (a flow-matching action head, ``System~1'') that advance at independent rates and communicate through a lightweight, non-blocking interface, as in dual-system designs~\cite{gr00tn1,helix,fisvla}.
Once the backbone leaves the control critical path, its latency no longer bottlenecks the action rate: the VLM can be scaled up for generalization at almost no cost to control frequency.
We call this \emph{latency-agnostic scaling}: backbone size no longer trades against reactivity, so model scale becomes a free design axis rather than a speed--quality compromise.
Decoupling also sharpens action quality on two axes: it removes most of the inference-to-execution delay, and it raises the action-update frequency so the fast stream can exploit sensory evidence that arrives far faster than a synchronous policy can emit actions.
Because the early actions of a predicted chunk are far more reliable than its autoregressively uncertain tail (Fig.~\ref{fig:intro}a), updating chunks more often keeps execution on the high-confidence portion of each prediction and curbs divergence between successive chunks (Fig.~\ref{fig:intro}b).

Asynchrony is not free: at deployment the action head must fuse \emph{stale} VLM semantics with \emph{fresh} perception, a distribution that synchronous training never presents.
We close this gap architecturally, with a fresh-vision encoder inside the action head, and in training, by exposing the policy to the desynchronized inputs it will meet at test time.

We propose \textbf{Decoupled Asynchronous System for Humanoids (DASH)} and evaluate it on a real humanoid with dexterous hands and in a RoboTwin simulator that we modify to run the environment and policy asynchronously, closer to real deployment.
As the backbone grows from $0.8$B to $8$B, the synchronous control rate falls from $9.7$ to $4.1$\,Hz, whereas DASH holds $20$\,Hz at every size, a $2.1$--$4.9\times$ higher control rate with higher success ($+1.0$ to $+5.9$\,pp); on the real humanoid, DASH succeeds in $23/24$ pick-and-place trials against $16/24$ for the synchronous pipeline.
Our contributions are as follows:
\begin{itemize}
    \item \textbf{DASH: an asynchronous VLA system deployed on a dexterous humanoid.} The policy is split into a slow VLM (System~2) and a fast flow-matching action head (System~1), run as three concurrent processes (environment, VLM server, action server) over the LCM communication protocol, so the control rate is set by the action head alone. Combined with an adaptive temporal aggregator that aligns overlapping chunks by fractional interpolation, DASH runs on a real humanoid for dexterous-hand manipulation with a fully specified, reproducible inference pipeline.
    \item \textbf{Latency-agnostic scaling.} Taking the backbone off the control critical path decouples control rate from backbone size. In an asynchronous RoboTwin benchmark that charges inference latency in simulated time, DASH holds a $50$\,ms action path and $20$\,Hz from $0.8$B to $8$B while the synchronous rate falls from $9.7$ to $4.1$\,Hz, with higher success at every size.
    \item \textbf{Cross-timescale fusion and a training recipe for asynchrony.} A frozen VLM plus a trainable action head with a ResNet vision encoder gives the policy fresh, adapted visual evidence to ground stale semantics, fusing cached multi-layer VLM states with fresh visual tokens from the head-clocked encoder inside the action head. Desynchronized training makes staleness awareness implicit: the VLM input is offset from the action-head input by a randomly sampled staleness, so the policy learns to act on stale context and fresh observations without any explicit staleness input, keeping training consistent with inference.
\end{itemize}

\section{Related Work}
\label{sec:related}

\subsection{Scaling and Adapting VLM Backbones}
Scaling the vision-language backbone improves generalization, but inference latency has kept deployed VLAs small: RT-2 showed that web-scale backbones transfer semantic knowledge to control, yet its largest models run far below the rates that reactive control requires~\cite{rt2}, so subsequent systems settle on compact backbones, e.g.\ 7B for OpenVLA and $2$--$3$B for $\pi_0$ and GR00T~N1~\cite{openvla,pi0,gr00tn1}.
How the backbone is adapted also divides existing recipes: RT-2 and OpenVLA fine-tune the full model~\cite{rt2,openvla}, $\pi_0$ trains an action expert jointly with the backbone~\cite{pi0}, GR00T~N1 trains a DiT head end-to-end while keeping the language model of its backbone frozen~\cite{gr00tn1}, HiRT adapts the VLM only through LoRA~\cite{hirt}, and DP-VLA uses an off-the-shelf VLM purely as a frozen feature extractor for its fast system~\cite{dpvla}.
In most of these, the backbone remains on the control-critical path, so its size is still bounded by the control rate.

\subsection{Dual-System Fast--Slow Architectures}
Slow--fast pairings are well established but are typically trained and deployed synchronously.
GR00T~N1 trains a DiT head on the intermediate features of a partially frozen VLM~\cite{gr00tn1}, Hi~Robot stacks a high-level VLM over $\pi_0$ through language~\cite{hirobot}, HiRT caches slow latents for a fast policy~\cite{hirt}, RoboDual and DP-VLA condition a specialist on generalist latents~\cite{robodual,dpvla}, FiS-VLA repurposes the last VLM blocks as its fast system~\cite{fisvla}, and Helix and DuoCore-FS run split systems at two rates on humanoid and whole-body platforms~\cite{helix,duocore}.
The interface between the two systems is rarely made explicit: how stale the slow system's output is when the fast system consumes it, how overlapping action chunks are aligned when inference latency varies, and which component lets the fast system act on a stale context are left unmeasured.

\subsection{Latency-Aware and Asynchronous Robot Control}
Existing treatments of inference latency intervene at inference time or at training time, and target chunk boundaries or state staleness rather than the slow system's features.
At inference time, RTC freezes committed actions and inpaints the remainder across chunk boundaries~\cite{rtc}, SmolVLA overlaps inference with execution through a server--client stack~\cite{smolvla}, and ACT-style temporal ensembling smooths overlapping chunks under a fixed, integer chunk offset~\cite{act}.
At training time, VLASH rolls robot states forward to the expected execution time~\cite{vlash}, TTAC conditions on executed action prefixes under simulated delays~\cite{ttac}, Helix applies a fixed offset between slow- and fast-system inputs calibrated to deployment latency~\cite{helix}, and TIDAL pairs stale semantic intent with fresh proprioception~\cite{tidal}; delay-aware reinforcement learning and streaming perception formalize acting under delay~\cite{delayrl,streamyolo}.
These methods target action or state staleness rather than the staleness of the slow system's features in a split policy.

\begin{figure*}[t]
\centering
\includegraphics[width=\textwidth]{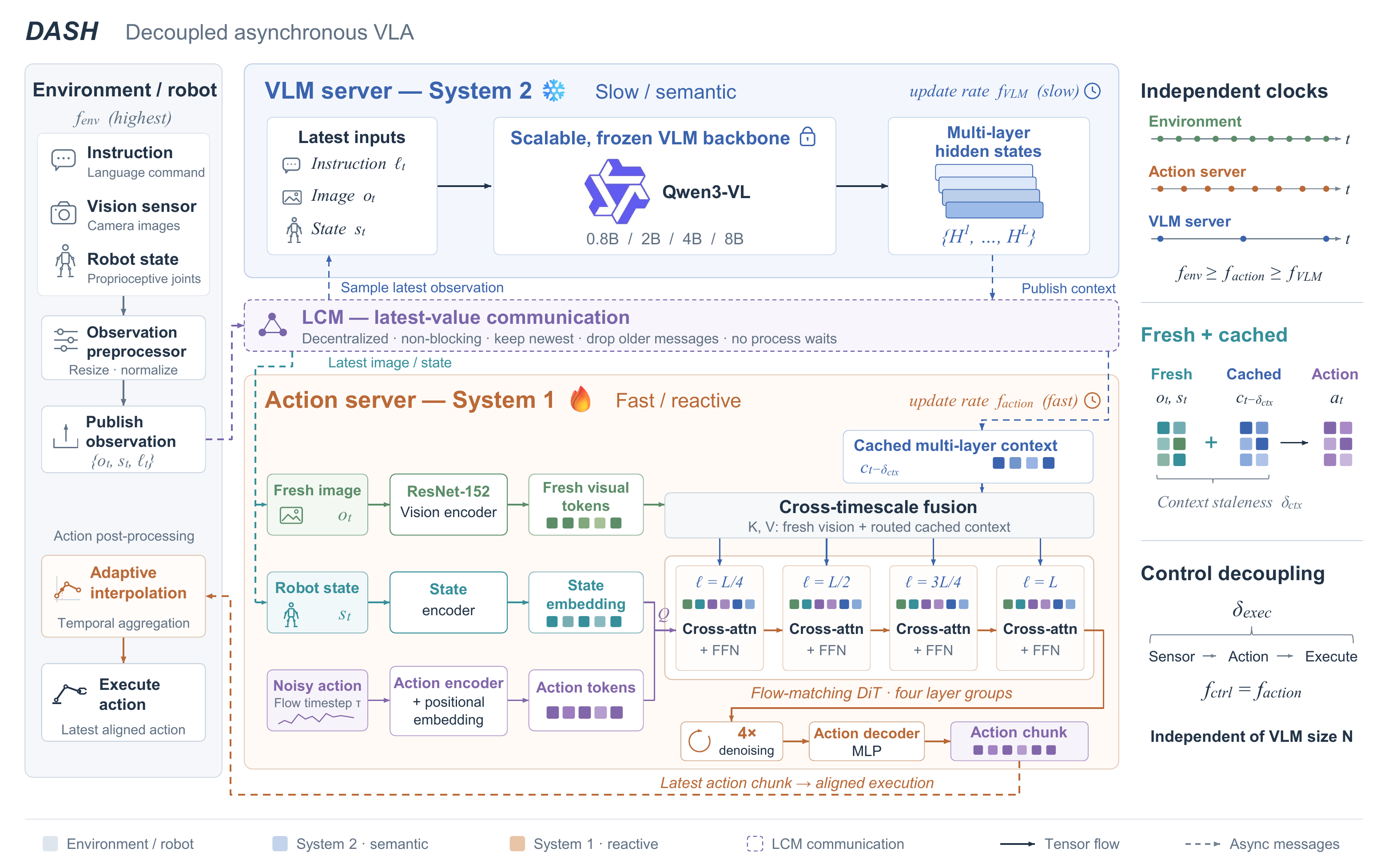}
\caption{DASH architecture. The policy is split into a slow VLM (System~2) and a fast action head (System~1), deployed as three concurrent processes (environment, VLM server, and action server) coupled by a decentralized, non-blocking latest-value bus (LCM). The VLM runs off the control loop and publishes a context (the hidden states of the routed VLM layers) that the action head reads with latest-value semantics, so the control rate is set by the action head alone.}
\label{fig:arch}
\end{figure*}

\section{Decoupled Asynchronous System for Humanoids (DASH)}
\subsection{Problem Formulation}
\label{subsec:formulation}

We study closed-loop manipulation in continuous time $t$: the environment and robot evolve with $t$, while onboard cameras and sensors expose an image $o_t$, a proprioceptive state $s_t$, and a language instruction $\ell_t$.
A standard VLA is a single end-to-end model that maps all three inputs to an action chunk of horizon $H$,
\begin{equation}
a_{t:t+H}=\pi\big(o_t,\,s_t,\,\ell_t\big).
\label{eq:monolithic}
\end{equation}

\noindent\textbf{Scaling is a latency trade-off.}
The model needs time $L$ to run, so a chunk computed from the inputs at $t$ is applied only at $t+L$, by which point the world has moved; we call this the \emph{inference--execution staleness} $\delta_{\text{exec}}\!=\!L$.
A larger backbone generalizes better and understands context more reliably, yet it is also slower since $L$ grows monotonically with model size $N$,
\begin{equation}
\delta_{\text{exec}}=L(N),\qquad \frac{\partial L}{\partial N}>0 .
\label{eq:scaling}
\end{equation}
Under this monolithic synchronous formulation, scaling up capability directly inflates staleness and degrades closed-loop control: capability and reactivity are in direct conflict.

\noindent\textbf{Asynchronous reformulation.}
We recast the single policy as two modules that may advance on independent clocks: a slow semantic module (the VLM, System~2) that produces a context $c_t$, and a fast reactive module (the action head, System~1) that emits the chunk,
\begin{equation}
c_t=\pi_{\text{VLM}}\big(o_t,s_t,\ell_t\big),\quad
a_{t:t+H}=\pi_{\text{action}}\big(s_t,o_t,c_{t-\Delta_{\text{VLM}}}\big).
\label{eq:async}
\end{equation}
Running asynchronously reshapes staleness in two ways.
First, $\delta_{\text{exec}}$ is now set by the fast action head alone, small and \emph{independent of the VLM size} $N$.
Second, the action head pairs a fresh observation with a context computed one VLM latency earlier, creating a \emph{context staleness}
\begin{equation}
\delta_{\text{ctx}}=\Delta_{\text{VLM}},
\label{eq:dctx}
\end{equation}
a temporal mismatch between the inputs of $\pi_{\text{VLM}}$ and $\pi_{\text{action}}$ that the monolithic model never faced.

\noindent\textbf{Objective.}
Our aim is to reduce both gaps: to drive $\delta_{\text{exec}}$ down and decouple it from $N$, so a large VLM no longer harms reactivity, and to make the action head robust to $\delta_{\text{ctx}}$, so that acting on a stale context does not cost accuracy.

\subsection{Decoupled Fast–Slow System for Latency-Agnostic Scaling}
\label{subsec:arch}
DASH splits the monolithic VLA policy into a fast action head (System~1) and a slow VLM (System~2), deployed as three concurrent processes: environment, VLM server, and action server (Fig.~\ref{fig:arch}).
Observation pre-processing and action post-processing execute in the environment process, so each server runs inference only.
The VLM turns the current image, state, and instruction into a context $c_t$ that the action head consumes; the contents and their fusion are specified in Sec.~\ref{subsec:fusion}.

The processes communicate over LCM~\cite{lcm}, a decentralized UDP publish/subscribe layer whose endpoints are matched by channel name without connection setup.
The environment is the publisher of observations and both servers are subscribers: it streams images and state continuously at the highest rate its sensors allow, without waiting for either server to finish.
Each subscriber buffers messages in a FIFO queue and, whenever a loop completes and needs new input, drains the queue to its most recent entry; no process blocks on a peer, and each conditions on the freshest available input.
As only preprocessed tensors are transmitted, per-message transport is at the millisecond scale and negligible against inference.
Execution thus follows a receding-horizon scheme driven by arrival rather than by a fixed schedule: the action server predicts a chunk of $H$ actions, the environment executes it from the entry aligned with the current time, discarding the prefix already overtaken by latency, and replaces it as soon as the next chunk arrives, with overlapping chunks blended by the aggregator of Sec.~\ref{subsec:runtime}.
Under DASH the horizon therefore slides forward once per action-head inference, independent of the VLM.
The three loops run independently at rates $f_{\text{env}}\!\ge\!f_{\text{action}}\!\ge\!f_{\text{VLM}}$, so a single context $c_{t-\Delta t}$ is reused across the action steps between two VLM updates, the runtime origin of $\delta_{\text{ctx}}$.

With the VLM off the control loop, the control rate equals the action-head rate and is independent of the VLM size $N$,
\begin{equation}
f_{\text{ctrl}}=f_{\text{action}}\ \perp\ N .
\label{eq:latency-agnostic}
\end{equation}
The backbone can therefore be frozen and enlarged as a drop-in replacement: increasing $N$ alters $c_t$ but leaves $f_{\text{ctrl}}$ and $\delta_{\text{exec}}$ fixed, decoupling capability from reactivity and resolving Eq.~\eqref{eq:scaling}.

\subsection{Cross-Timescale Fusion via an Action-Head Vision Encoder}
\label{subsec:fusion}

Running the two systems at independent rates means the action head must combine information from two time scales: semantics carried by a VLM context that is stale, and perception of the scene that is fresh.
DASH fuses them inside the action head, injecting the routed VLM hidden states together with fresh visual tokens supplied by a ResNet action-head vision encoder that runs on the head's own clock.

We instantiate the VLM with a pretrained Qwen3-VL backbone of $L_{\text{vlm}}$ layers.
The action head is a flow-matching DiT of $L_{\text{dit}}\!=\!12$ \emph{cross-only} blocks, each applying one cross-attention followed by an FFN without a dedicated self-attention layer, with hidden width $D\!=\!1024$, $8$ heads, and $K\!=\!4$ integration steps at inference.
Its queries $Z^{\text{sa}}_t\in\mathbb{R}^{(1+H)\times D}$ are the robot state, embedded by a state encoder $E_s$, and the noised action chunk $a^{\tau}_{t:t+H}$, embedded jointly with the flow-matching timestep $\tau$ by an action encoder $E_a$ and given a learned positional embedding.
A cross-layer mapping $\phi$ selects which VLM layers feed which DiT blocks; the VLM server publishes the hidden states $H^{(\ell)}$ of these routed layers, and the action head reads the most recent entry at every inference step, so the same states serve consecutive action steps until the next VLM update arrives.
For block $d=0,\dots,L_{\text{dit}}-1$,
\begin{equation}
\phi(d)=\Big\lceil \tfrac{\lfloor 4d/L_{\text{dit}}\rfloor+1}{4}\,L_{\text{vlm}} \Big\rceil ,
\label{eq:phi}
\end{equation}
i.e.\ four contiguous bands of blocks read layers at $25/50/75/100\%$ of $L_{\text{vlm}}$, so early blocks receive shallow features and late blocks semantic ones; the mapping rescales with $L_{\text{vlm}}$ and thus transfers across backbone sizes.
In parallel, a ResNet-152 action-head vision encoder $E_v$ runs on the head's own current image, $V^{\text{fresh}}_t\!=\!E_v(o_t)$.
With projectors $W_{\phi(d)},W_v$ mapping both streams to width $D$, block $d$ attends over the concatenation of routed stale context and fresh visual tokens,
\begin{align}
K_d=V_d&=\big[\,\underbrace{W_{\phi(d)}H^{(\phi(d))}_{t-\Delta t}}_{\text{stale context}}\ ;\ \underbrace{W_v V^{\text{fresh}}_{t}}_{\text{fresh vision}}\,\big],
\label{eq:kv}\\[2pt]
Z^{\text{sa}}_{d+1}&=\mathrm{FFN}\big(\mathrm{CrossAttn}(Q{=}Z^{\text{sa}}_{d},K_d,V_d)\big),
\label{eq:fusion}
\end{align}
and an action decoder maps $Z^{\text{sa}}_{L_{\text{dit}}}$ to the predicted velocity field (Fig.~\ref{fig:arch}).
Since $c_{t-\Delta t}$ is constant within a chunk, the context term is computed once and reused across all $K$ steps, while $V^{\text{fresh}}_t$ refreshes on the action-head clock.

\begin{figure}[t]
\centering
\includegraphics[width=\linewidth]{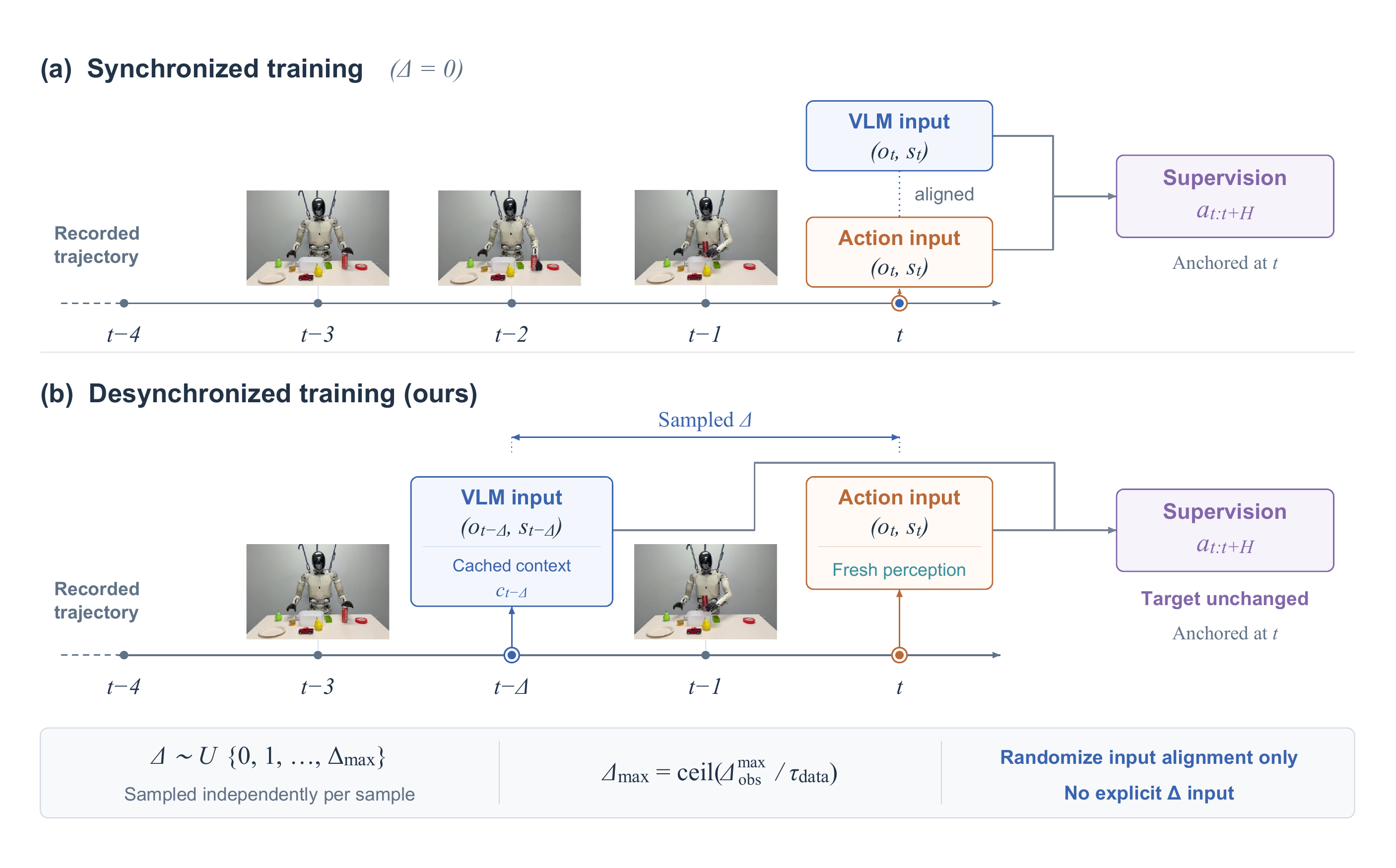}
\caption{Desynchronized training data. The two visual channels are sampled at different times so the training inputs reproduce the staleness met at deployment, while the supervision target remains anchored at the action head's timestamp.}
\label{fig:desync}
\end{figure}

\noindent\textbf{Training recipe for latency-agnostic scaling.}
We freeze the VLM backbone and train only the DiT and its ResNet vision encoder.
Freezing preserves the grounding and instruction-following acquired at web scale, and it fixes the trainable parameter count, so enlarging the backbone leaves the training cost unchanged, the training-side counterpart of Eq.~\eqref{eq:latency-agnostic}.
Freezing alone, however, is insufficient: the backbone's vision encoder is frozen with it and its features reach the head only after $\Delta t$, so the policy would act on visual evidence that is both non-adapted and stale, which degrades closed-loop success.
The ResNet encoder removes both deficiencies at once: it adapts to the manipulation distribution while the semantic pathway stays frozen, and it runs on the action-head clock, so the evidence entering Eq.~\eqref{eq:kv} refreshes at $f_{\text{action}}$ rather than $f_{\text{VLM}}$, at a latency that leaves $\delta_{\text{exec}}$ unaffected.

\subsection{Desynchronized Training for Train–Inference Consistency}
\label{subsec:train}

At deployment the action head always pairs a fresh image and state with a context produced one VLM cycle earlier (Eq.~\eqref{eq:async}), whereas standard training pairs all inputs at a single timestamp, $\Delta t{=}0$.
A policy fitted only on aligned tuples is never asked to reconcile the two time scales, so at test time it meets an input distribution it has never seen.
We close this gap by reproducing the deployment misalignment in the training data.

\noindent\textbf{Desynchronized data construction.}
In the pipeline of Sec.~\ref{subsec:arch} the VLM and the action head consume the same camera stream on different clocks, the VLM at low rate and the action head at high rate.
Training thus needs only to sample the two inputs at different times, leaving every interface unchanged.
Given a reference time $t$, we keep the action head on the current frame and draw the VLM branch from an earlier one,
\begin{equation}
\big(o^{\text{vlm}},s^{\text{vlm}}\big)=\big(o_{t-\Delta},\,s_{t-\Delta}\big),\qquad
\big(o^{\text{act}},s^{\text{act}}\big)=\big(o_{t},\,s_{t}\big),
\label{eq:desync-sample}
\end{equation}
where $\Delta$ counts image ticks of the recorded stream and is drawn independently per sample from a uniform distribution
\begin{equation}
\Delta \sim \mathcal{U}\{0,\,1,\ldots,\,\Delta_{\max}\},\qquad
\Delta_{\max}=\Big\lceil \tfrac{\Delta^{\max}_{\text{obs}}}{\tau_{\text{data}}} \Big\rceil .
\label{eq:delta-sampling}
\end{equation}
\begin{figure}[t]
\centering
\includegraphics[width=\linewidth]{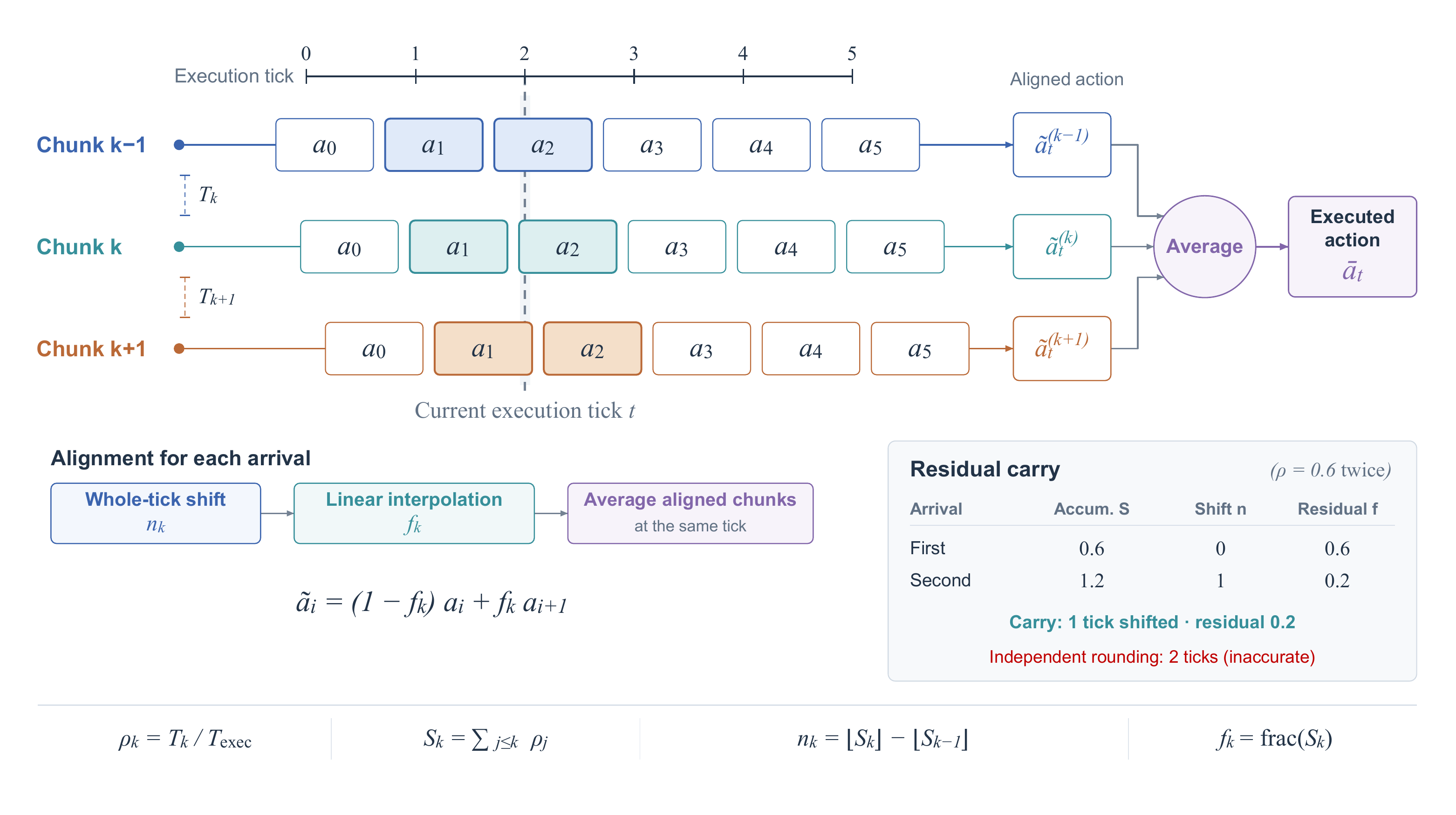}
\caption{Adaptive interpolation for temporal aggregation. The fractional offset measured from the true chunk arrival interval is split into a whole-tick shift and a linear interpolation between adjacent entries, with the residual fractions accumulated so the alignment tracks elapsed execution time.}
\label{fig:aggregator}
\end{figure}
Here $\Delta^{\max}_{\text{obs}}$ is the largest observation time difference between the two channels measured at deployment, and $\tau_{\text{data}}$ is the sampling period of the training data, so the ratio converts a wall-clock latency into a number of image ticks.
The lower bound $\Delta{=}0$ recovers the synchronous case and the upper bound matches the worst case the runtime can produce, so the training support covers the staleness actually encountered.
The supervision target is unchanged: the flow-matching loss is still taken on the chunk $a_{t:t+H}$ anchored at the action head's timestamp, now conditioned on the stale context $c_{t-\Delta}$, so only the input alignment is randomized and the recipe composes with any backbone or action head.
Sampling $\Delta$ per example rather than fixing it covers the whole staleness range, since the runtime offset varies with load and backbone size.

\noindent\textbf{Implicit staleness.}
We provide no explicit staleness input, auxiliary token, or gating branch: the training distribution alone carries the information, and the policy learns from the data to weigh fresh perception against stale semantics.
The deployed model is therefore identical to the synchronous one, adds no inference cost, and its training and inference input distributions coincide.

\subsection{Adaptive Temporal Aggregation under Asynchronous Inference}
\label{subsec:runtime}

\begin{table*}[t]
\centering
\caption{Backbone scaling, synchronous inference vs.\ DASH (50 tasks, 250 episodes). Async (w/o desynced training): the DASH runtime with aligned training. $\Delta$ = DASH $-$ Sync (pp).}
\label{tab:matrix}
\begin{tabular}{lcccccccccc}
\toprule
 & \multicolumn{2}{c}{Latency (ms)} & \multicolumn{3}{c}{Control rate (Hz)} & \multicolumn{4}{c}{Success rate (\%)} \\
\cmidrule(lr){2-3}\cmidrule(lr){4-6}\cmidrule(lr){7-10}
Backbone & Sync & DASH & Sync & DASH & Speedup & Sync & Async (w/o desynced training) & DASH & $\Delta$ \\
\midrule
0.8B & 103 & \textbf{50} & 9.7 & \textbf{20} & 2.1$\times$ & 64.5 & 61.8 & \textbf{65.5} & $+$1.0 \\
2B   & 132 & \textbf{50} & 7.6 & \textbf{20} & 2.6$\times$ & 66.4 & 65.5 & \textbf{69.1} & $+$2.7 \\
4B   & 183 & \textbf{50} & 5.5 & \textbf{20} & 3.6$\times$ & 63.6 & 62.3 & \textbf{69.5} & $+$5.9 \\
8B   & 243 & \textbf{50} & 4.1 & \textbf{20} & 4.9$\times$ & 62.7 & 62.7 & \textbf{65.9} & $+$3.2 \\
\bottomrule
\end{tabular}
\end{table*}

\begin{table}[t]
\centering
\caption{Measured inference latency (ms, RTX~4090).}
\label{tab:latency}
\setlength{\tabcolsep}{4pt}
\begin{tabular}{lc}
\toprule
Component & Latency (ms) \\
\midrule
VLM, Qwen3.5-0.8B & 53 \\
VLM, Qwen3-VL-2B  & 82 \\
VLM, Qwen3-VL-4B  & 133 \\
VLM, Qwen3-VL-8B  & 193 \\
\midrule
Image preprocessing & 15 \\
Action head (ResNet-152 $+$ 4-step flow-matching DiT) & 35 \\
\bottomrule
\end{tabular}
\end{table}

Consecutive chunks overlap, so several of them predict the same control tick, and a temporal aggregator averages those predictions to suppress noise and smooth the executed motion~\cite{act,rtc}.
Averaging first requires alignment: a chunk received $\rho$ ticks ago must be shifted by $\rho$ before its entries refer to the current tick, where $\rho$ is the inference interval in units of the tick execution period $T_{\text{exec}}$.

Synchronous inference is gated to a fixed period, so $\rho$ is a known integer and the shift is exact.
Asynchrony removes this guarantee: inference is no longer gated and its duration varies per chunk, making $\rho$ a time-varying real number whose rounding over-advances the buffer, which is what makes naive asynchronous execution fast and jerky.

We instead measure the arrival interval $T_k$ between the $k$-th chunk and its predecessor and use the exact, fractional offset
\begin{equation}
\rho_k=T_k/T_{\text{exec}} .
\label{eq:adaptive}
\end{equation}
A chunk must then be shifted by a non-integer amount, which we realize in two parts.
The integer part selects how many whole ticks to shift, and the fraction $f_k$ resamples between adjacent entries by linear interpolation, $\tilde{a}=(1-f_k)\,a_{i}+f_k\,a_{i+1}$, applied per action dimension.
Discarding the fraction at each chunk would let the alignment drift, so we accumulate the offsets and take the shift as the increment of their running total,
\begin{equation}
n_k=\Big\lfloor \textstyle\sum_{j\le k}\rho_j \Big\rfloor-\Big\lfloor \textstyle\sum_{j<k}\rho_j \Big\rfloor,
\qquad
f_k=\Big\{\textstyle\sum_{j\le k}\rho_j\Big\},
\label{eq:carry}
\end{equation}
with $\{\cdot\}$ the fractional part, so fractions carry into a whole-tick shift once they sum past one: two successive chunks with $\rho\!=\!0.6$ leave the buffer unshifted and interpolated at $0.6$, then shift it by one tick and interpolate at $0.2$.
The ensemble therefore follows the true inference cadence, keeping execution at the pace of the training data and turning the higher chunk density of asynchronous inference into finer motion rather than jitter.

\section{Experiments}
\label{sec:exp}

\begin{figure}[t]
\centering
\begin{tikzpicture}
\begin{axis}[
    width=0.95\linewidth, height=4.0cm,
    ybar, bar width=8pt,
    xlabel={VLM backbone size},
    ylabel={Success rate (\%)},
    ymin=50, ymax=78,
    symbolic x coords={0.8B,2B,4B,8B},
    xtick=data,
    legend style={at={(0.5,1.02)}, anchor=south, legend columns=-1, font=\scriptsize},
    ymajorgrids,
    enlarge x limits=0.18,
    nodes near coords,
    every node near coord/.append style={font=\tiny, black, rotate=90, anchor=west, /pgf/number format/.cd, fixed, precision=1, zerofill},
]
\addplot[fill=blue!35, draw=blue!60!black] coordinates {(0.8B,64.5) (2B,66.4) (4B,63.6) (8B,62.7)};
\addlegendentry{Sync}
\addplot[fill=gray!45, draw=gray!70!black] coordinates {(0.8B,61.8) (2B,65.5) (4B,62.3) (8B,62.7)};
\addlegendentry{Async (w/o desynced training)}
\addplot[fill=red!40, draw=red!60!black] coordinates {(0.8B,65.5) (2B,69.1) (4B,69.5) (8B,65.9)};
\addlegendentry{DASH}
\end{axis}
\end{tikzpicture}
\caption{Success rate by backbone size (Table~\ref{tab:matrix}): synchronous inference, Async (w/o desynced training), and DASH. DASH is higher than synchronous inference at every size ($+1.0/{+}2.7/{+}5.9/{+}3.2$\,pp) while its action path stays at $50$\,ms.}
\label{fig:sr}
\end{figure}
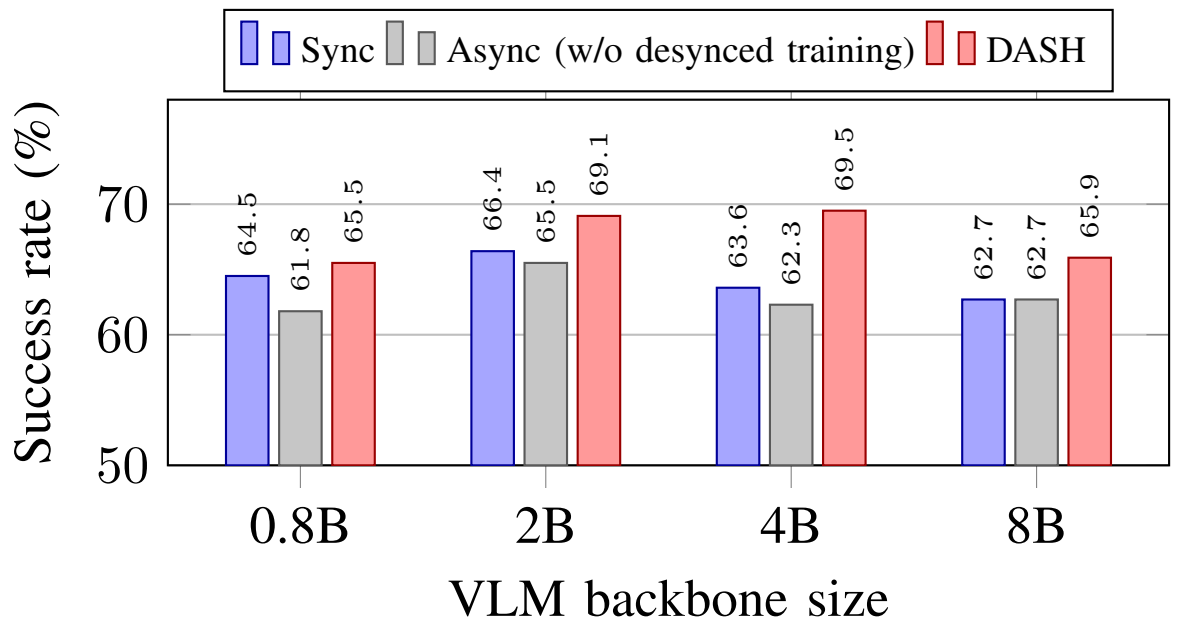

\begin{figure*}[t]
\centering
\includegraphics[width=\textwidth]{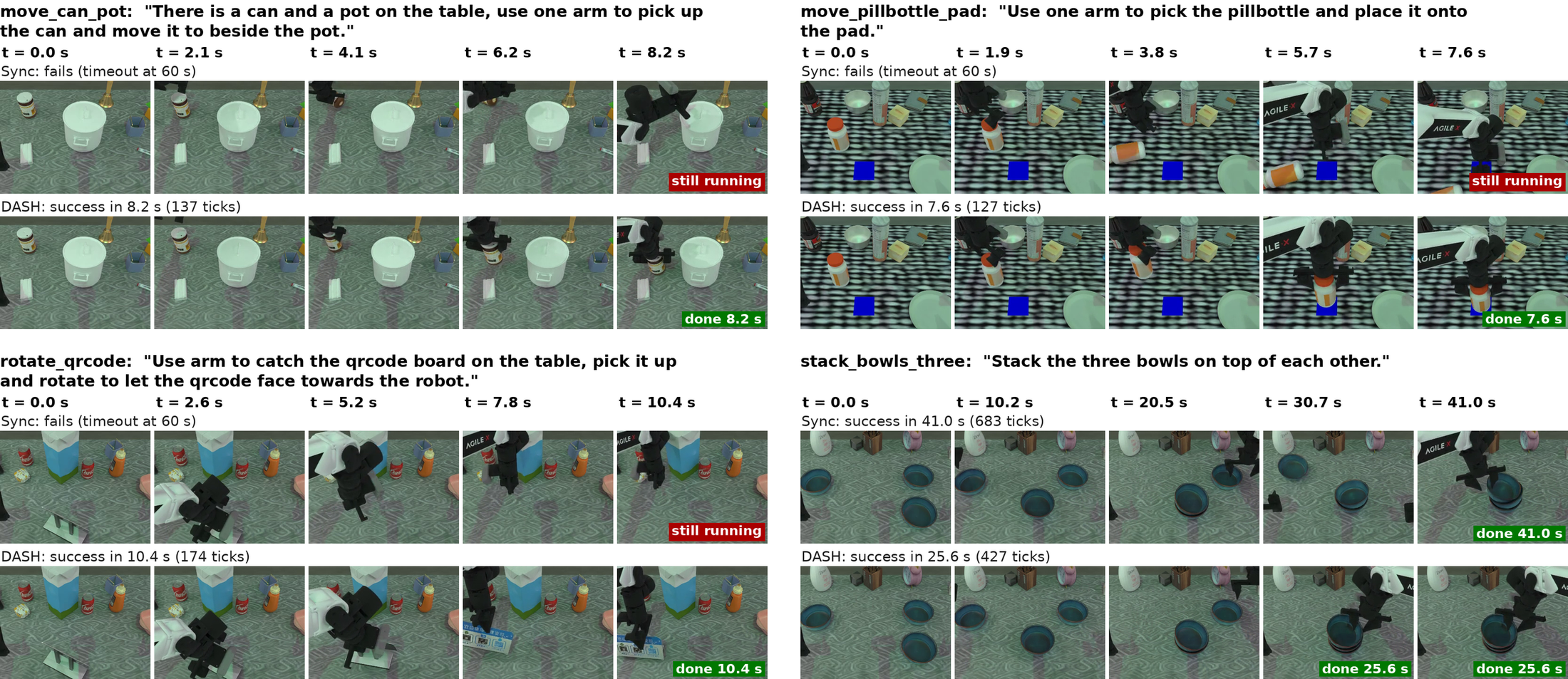}
\caption{Matched simulation episodes, synchronous inference (top) vs.\ DASH (bottom). Frames are sampled at equal intervals of simulated time up to the completion time of the successful run; \emph{done} marks completion, and \emph{still running} marks the run that later times out.}
\label{fig:simqual}
\end{figure*}

\begin{figure*}[t]
\centering
\includegraphics[width=\textwidth]{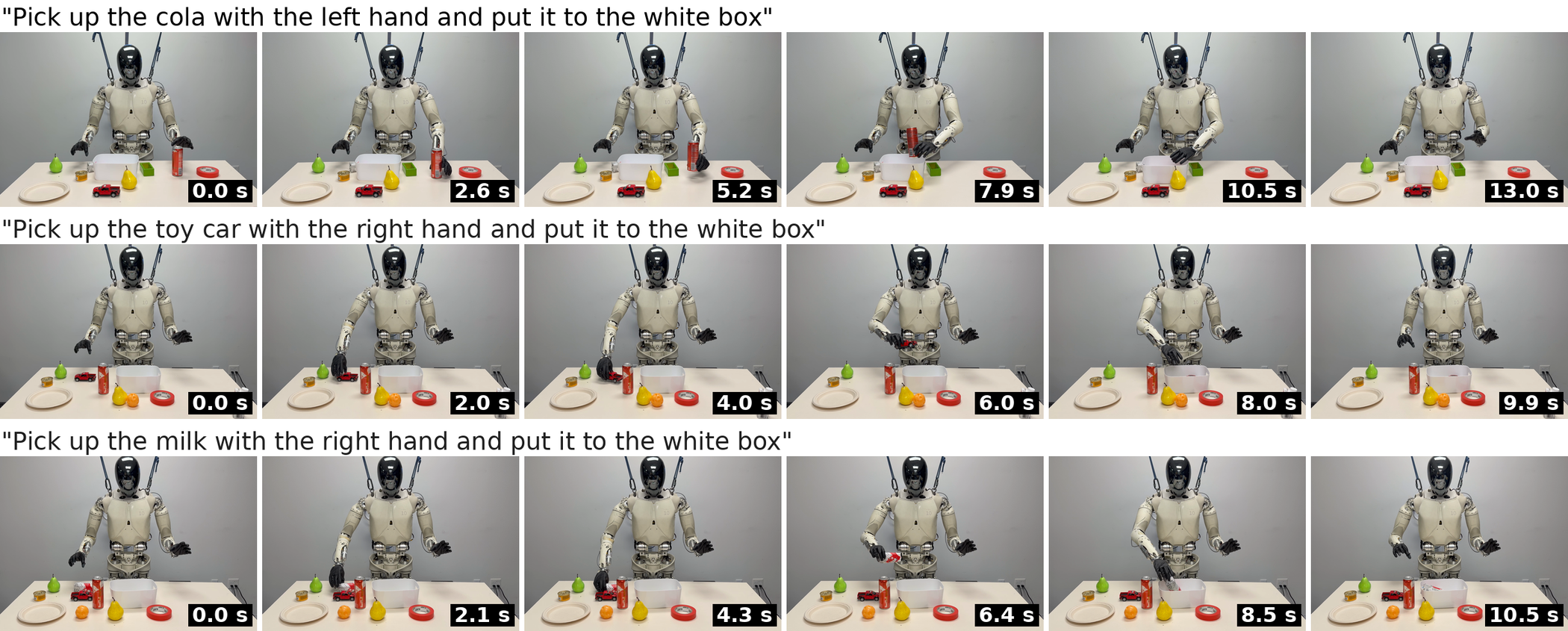}
\caption{DASH on the real humanoid: three of the language-conditioned pick-and-place instructions, six frames each at equal intervals of wall-clock time (time stamps in the corners).}
\label{fig:real}
\end{figure*}

\subsection{Experimental Setup}
\label{sec:exp-setup}

\noindent\textbf{Asynchronous simulation.}
We evaluate closed loop in RoboTwin (aloha embodiment).
Native RoboTwin pauses the world while the policy runs, so inference latency never affects the outcome.
We make it asynchronous by charging latency in simulated time: the world advances on a fixed control tick, and an inference result computed from the observation at $t$ becomes available only at $t+\text{latency}$, while the robot keeps executing its previous chunk.
In the split configuration the action head likewise pairs each fresh observation with the newest available VLM context, so $\delta_{\text{exec}}$ and $\delta_{\text{ctx}}$ (Sec.~\ref{subsec:formulation}) arise from the schedule itself.

\noindent\textbf{Real humanoid.}
We deploy the wall-clock asynchronous runtime (Sec.~\ref{subsec:arch}) with the adaptive temporal aggregator (Sec.~\ref{subsec:runtime}) on a humanoid robot with two dexterous hands.
The tasks are language-conditioned tabletop pick and place: each instruction names the object, the hand to use, and the target, e.g.\ ``pick up the toy car with the right hand and put it to the plate'' or ``pick up the milk with the right hand and put it to the white box'', over grocery and household objects (a cola can, a bag of chips, a toy car, a milk carton).
Each configuration runs $24$ trials.

\noindent\textbf{Models and training.}
The backbone is Qwen3.5-0.8B or Qwen3-VL-2B/4B/8B; the action head is the 12-layer, 1024-wide DiT with ResNet-152 and 4-step flow matching.
Each backbone is trained both with aligned data and with desynchronized data (Sec.~\ref{subsec:train}), where the VLM branch is offset by a number of frames derived from the deployed context delay; Table~\ref{tab:matrix} reports the asynchronous runtime under both recipes.

\noindent\textbf{Simulation tasks and metric.}
We use the 50-task RoboTwin suite, which covers pick-and-place and relocation, articulated objects and buttons, tool use and in-hand reorientation, and stacking, sorting, and handover.
Each configuration is evaluated over the full suite with five episodes per task, each from a different scene seed, for 250 episodes per configuration.
Part of the suite is out-of-distribution with respect to the training demonstrations (tasks unseen in training), which keeps absolute success rates moderate.
On both platforms we report success rate (SR), the fraction of episodes or trials that complete the task within the budget.
In simulation, SR is computed over valid episodes with successful initialization.

\subsection{Simulation Evaluation for Latency-Agnostic Scaling}
\label{sec:exp-scaling}

Table~\ref{tab:matrix} and Fig.~\ref{fig:sr} compare three configurations on the 50-task suite: synchronous inference (monolithic; actions wait for VLM$+$head) and the asynchronous runtime without desynchronized training (split; actions wait only for the fast path), which share one checkpoint trained on aligned data, and DASH, the same split runtime with the checkpoint trained on desynchronized data (Sec.~\ref{subsec:train}).

\noindent\textbf{Latency and reactivity.}
Table~\ref{tab:latency} reports the measured latency of each component on an RTX~4090 at deployment resolution ($480{\times}640$): VLM latency grows from $53$ to $193$\,ms with backbone size, whereas image preprocessing ($15$\,ms) and the action head ($35$\,ms) do not depend on it.
A synchronous inference therefore takes $103/132/183/243$\,ms for 0.8B/2B/4B/8B, cutting the control rate from $9.7$ to $4.1$\,Hz, whereas the DASH action path takes $50$\,ms at every size and runs at $20$\,Hz: $2.1\times$ the synchronous rate at 0.8B and $4.9\times$ at 8B.
Each tick therefore executes from the freshest chunk, whereas the synchronous pipeline executes ever longer and staler portions of each chunk as the backbone grows.

\noindent\textbf{Success.}
DASH exceeds synchronous inference at every size ($\Delta=+1.0/{+}2.7/{+}5.9/{+}3.2$\,pp for 0.8B/2B/4B/8B), with the largest gain and the highest success ($69.5\%$) at 4B, the sweet spot of this suite.
The Async (w/o desynced training) column isolates the contribution of desynchronized training: the same asynchronous runtime with aligned training sits slightly below synchronous inference, so asynchrony alone does not buy success; training the head on the staleness it meets at deployment lifts it by $+3.2$ to $+7.2$\,pp over the aligned asynchronous configuration.
Fig.~\ref{fig:simqual} shows four matched episodes in which synchronous inference stalls or times out while DASH completes the task.

\subsection{Real-World Evaluation on a Dexterous Humanoid}
\label{sec:exp-real}

Table~\ref{tab:realrobot} compares DASH with the adaptive aggregator against synchronous inference with a fixed-offset aggregator on the real humanoid: $23/24$ successful trials against $16/24$.
Figs.~\ref{fig:teaser} and~\ref{fig:real} show DASH executing the instructions.

The gap traces to chunk alignment (Sec.~\ref{subsec:runtime}): a fixed offset misaligns chunk boundaries whenever the true inference interval deviates from the assumed one, producing jitter at chunk transitions and, under integer ceiling, motion faster than the demonstrations but less accurate in grasp placement; adaptive alignment removes both.
DASH runs stably on hardware with no packet loss and is reactive under perturbation.

\begin{table}[t]
\centering
\caption{Real-robot pick and place ($24$ trials per configuration).}
\label{tab:realrobot}
\begin{tabular}{lcc}
\toprule
Method & Success & SR (\%) \\
\midrule
Sync, fixed-offset aggregator & 16/24 & 66.7 \\
DASH, adaptive aggregator (ours) & \textbf{23/24} & \textbf{95.8} \\
\bottomrule
\end{tabular}
\end{table}

\subsection{Recipe Ablation: Freezing and Fresh Vision}
\label{sec:exp-recipe}

Table~\ref{tab:freezing} ablates the training recipe of Sec.~\ref{subsec:fusion} at 4B under asynchronous inference with identical latency accounting on the 50-task suite.

\begin{table}[t]
\centering
\caption{Recipe ablation under asynchronous inference (4B, 50 tasks, 250 episodes).}
\label{tab:freezing}
\begin{tabular}{lc}
\toprule
Recipe & Success rate (\%) \\
\midrule
full fine-tune, no encoder & 62.6 \\
frozen, no encoder         & 55.0 \\
frozen $+$ ResNet-152 (ours) & \textbf{69.5} \\
\bottomrule
\end{tabular}
\end{table}

As argued in Sec.~\ref{subsec:fusion}, freezing the backbone without a head encoder leaves System~1 with visual features that are neither adapted nor fresh; success drops to $55.0\%$, $7.6$\,pp below the full fine-tune baseline.
\textbf{Unfreezing is not the fix}: full fine-tuning without an encoder recovers only to $62.6\%$: the adapted features still reach the head stale, and the recipe gives up the frozen backbone's web-scale grounding and fixed training cost (Sec.~\ref{subsec:fusion}).
\textbf{Fresh vision is the fix}: adding the ResNet encoder to the frozen backbone lifts success to $69.5\%$, $+14.5$\,pp over the frozen baseline and $+6.9$\,pp over full fine-tuning, while training only the head.

\section{Conclusion}

We presented DASH, a decoupled asynchronous VLA system that takes the vision-language backbone off the control-critical path, so the control rate no longer depends on backbone size, while cross-timescale fusion, desynchronized training, and an adaptive temporal aggregator let the action head act on stale semantics without loss.
DASH holds a constant control rate across backbones from 0.8B to 8B with higher success than synchronous inference, in simulation and on a real dexterous humanoid.
The tasks evaluated here are quasi-static; we expect the constant control rate and fresh perception of DASH to matter most on dynamic and long-horizon tasks, where a synchronous pipeline must pay for backbone capacity with latency, and extending the evaluation to these settings is our next step.





\bibliographystyle{IEEEtran}
\bibliography{references}

@inproceedings{rt2,
  author    = {Brohan, Anthony and Brown, Noah and Carbajal, Justice and Chebotar, Yevgen and Chen, Xi and Choromanski, Krzysztof and Ding, Tianli and Driess, Danny and Dubey, Avinava and Finn, Chelsea and others},
  title     = {{RT-2}: Vision-Language-Action Models Transfer Web Knowledge to Robotic Control},
  booktitle = {Conference on Robot Learning (CoRL)},
  year      = {2023},
  note      = {arXiv:2307.15818}
}

@article{openvla,
  author  = {Kim, Moo Jin and Pertsch, Karl and Karamcheti, Siddharth and Xiao, Ted and Balakrishna, Ashwin and Nair, Suraj and Rafailov, Rafael and Foster, Ethan and Lam, Grace and Sanketi, Pannag and others},
  title   = {{OpenVLA}: An Open-Source Vision-Language-Action Model},
  journal = {arXiv preprint arXiv:2406.09246},
  year    = {2024}
}

@inproceedings{pi0,
  author    = {Black, Kevin and Brown, Noah and Driess, Danny and Esmail, Adnan and Equi, Michael and Finn, Chelsea and Fusai, Niccolo and Groom, Lachy and Hausman, Karol and Ichter, Brian and others},
  title     = {$\pi_0$: A Vision-Language-Action Flow Model for General Robot Control},
  booktitle = {Robotics: Science and Systems (RSS)},
  year      = {2025},
  note      = {arXiv:2410.24164}
}

@article{gr00tn1,
  author  = {{NVIDIA} and Bjorck, Johan and Casta{\~n}eda, Fernando and Cherniadev, Nikita and Da, Xingye and Ding, Runyu and Fan, Linxi and Fang, Yu and Fox, Dieter and Hu, Fengyuan and Huang, Spencer and others},
  title   = {{GR00T~N1}: An Open Foundation Model for Generalist Humanoid Robots},
  journal = {arXiv preprint arXiv:2503.14734},
  year    = {2025}
}

@inproceedings{hirobot,
  author    = {Shi, Lucy Xiaoyang and Ichter, Brian and Equi, Michael and Ke, Liyiming and Pertsch, Karl and Vuong, Quan and Tanner, James and Walling, Anna and Wang, Haohuan and Fusai, Niccolo and others},
  title     = {{Hi Robot}: Open-Ended Instruction Following with Hierarchical Vision-Language-Action Models},
  booktitle = {International Conference on Machine Learning (ICML)},
  year      = {2025},
  note      = {arXiv:2502.19417}
}

@misc{helix,
  author       = {{Figure AI}},
  title        = {Helix: A Vision-Language-Action Model for Generalist Humanoid Control},
  howpublished = {\url{https://www.figure.ai/news/helix}},
  month        = feb,
  year         = {2025}
}

@article{hirt,
  author  = {Zhang, Jianke and Guo, Yanjiang and Chen, Xiaoyu and Wang, Yen-Jen and Hu, Yucheng and Shi, Chengming and Chen, Jianyu},
  title   = {{HiRT}: Enhancing Robotic Control with Hierarchical Robot Transformers},
  journal = {arXiv preprint arXiv:2410.05273},
  year    = {2024}
}

@article{robodual,
  author  = {Bu, Qingwen and Li, Hongyang and Chen, Li and Cai, Jisong and Zeng, Jia and Cui, Heming and Yao, Maoqing and Qiao, Yu},
  title   = {Towards Synergistic, Generalized, and Efficient Dual-System for Robotic Manipulation},
  journal = {arXiv preprint arXiv:2410.08001},
  year    = {2024}
}

@article{dpvla,
  author  = {Han, ByungOk and Kim, Jaehong and Jang, Jinhyeok},
  title   = {A Dual Process {VLA}: Efficient Robotic Manipulation Leveraging {VLM}},
  journal = {arXiv preprint arXiv:2410.15549},
  year    = {2024}
}

@article{fisvla,
  author  = {Chen, Hao and Liu, Jiaming and Gu, Chenyang and Liu, Zhuoyang and Zhang, Renrui and Li, Xiaoqi and He, Xiao and Guo, Yandong and Fu, Chi-Wing and Zhang, Shanghang and Heng, Pheng-Ann},
  title   = {Fast-in-Slow: A Dual-System Foundation Model Unifying Fast Manipulation within Slow Reasoning},
  journal = {arXiv preprint arXiv:2506.01953},
  year    = {2025}
}

@article{smolvla,
  author  = {Shukor, Mustafa and Aubakirova, Dana and Capuano, Francesco and Kooijmans, Pepijn and Palma, Steven and Zouitine, Adil and Aractingi, Michel and Pascal, Caroline and Russi, Martino and Marafioti, Andres and others},
  title   = {{SmolVLA}: A Vision-Language-Action Model for Affordable and Efficient Robotics},
  journal = {arXiv preprint arXiv:2506.01844},
  year    = {2025}
}

@article{duocore,
  author  = {{Astribot Team}},
  title   = {Asynchronous Fast-Slow Vision-Language-Action Policies for Whole-Body Robotic Manipulation},
  journal = {arXiv preprint arXiv:2512.20188},
  year    = {2025}
}

@inproceedings{rtc,
  author    = {Black, Kevin and Galliker, Manuel Y. and Levine, Sergey},
  title     = {Real-Time Execution of Action Chunking Flow Policies},
  booktitle = {Advances in Neural Information Processing Systems (NeurIPS)},
  year      = {2025},
  note      = {arXiv:2506.07339}
}

@article{vlash,
  author  = {Tang, Jiaming and Sun, Yufei and Zhao, Yilong and Yang, Shang and Lin, Yujun and Zhang, Zhuoyang and Hou, James and Lu, Yao and Liu, Zhijian and Han, Song},
  title   = {{VLASH}: Real-Time {VLAs} via Future-State-Aware Asynchronous Inference},
  journal = {arXiv preprint arXiv:2512.01031},
  year    = {2025}
}

@article{ttac,
  author  = {Black, Kevin and Ren, Allen Z. and Equi, Michael and Levine, Sergey},
  title   = {Training-Time Action Conditioning for Efficient Real-Time Chunking},
  journal = {arXiv preprint arXiv:2512.05964},
  year    = {2025}
}

@article{tidal,
  author  = {Sun, Yuteng and Wang, Haoran and Bai, Ruofei and Li, Zhengguo and Li, Jun and Chuah, Meng Yee and Yau, Wei Yun},
  title   = {{TIDAL}: Temporally Interleaved Diffusion and Action Loop for High-Frequency {VLA} Control},
  journal = {arXiv preprint arXiv:2601.14945},
  year    = {2026}
}

@inproceedings{act,
  author    = {Zhao, Tony Z. and Kumar, Vikash and Levine, Sergey and Finn, Chelsea},
  title     = {Learning Fine-Grained Bimanual Manipulation with Low-Cost Hardware},
  booktitle = {Robotics: Science and Systems (RSS)},
  year      = {2023},
  note      = {arXiv:2304.13705}
}

@inproceedings{lcm,
  author    = {Huang, Albert S. and Olson, Edwin and Moore, David C.},
  title     = {{LCM}: Lightweight Communications and Marshalling},
  booktitle = {IEEE/RSJ International Conference on Intelligent Robots and Systems (IROS)},
  year      = {2010}
}

@inproceedings{delayrl,
  author    = {Bouteiller, Yann and Ramstedt, Simon and Beltrame, Giovanni and Pal, Christopher and Binas, Jonathan},
  title     = {Reinforcement Learning with Random Delays},
  booktitle = {International Conference on Learning Representations (ICLR)},
  year      = {2021}
}

@inproceedings{streamyolo,
  author    = {Yang, Jinrong and Liu, Songtao and Li, Zeming and Li, Xiaoping and Sun, Jian},
  title     = {Real-Time Object Detection for Streaming Perception},
  booktitle = {IEEE/CVF Conference on Computer Vision and Pattern Recognition (CVPR)},
  year      = {2022}
}

\end{document}